\documentclass{article}

\usepackage{microtype}
\usepackage{graphicx}
\usepackage{subfigure}
\usepackage{booktabs} %

\usepackage{hyperref}
\usepackage[a4paper, total={5.5in, 9in}]{geometry}

\usepackage{amssymb}%
\usepackage{pifont}
\usepackage{xspace}
\newcommand{\cmark}{\ding{51}\xspace}%
\newcommand{\xmark}{\ding{55}\xspace}%

\usepackage[utf8]{inputenc} %
\usepackage[T1]{fontenc}    %
\usepackage{hyperref}       %
\usepackage{url}            %
\usepackage{booktabs}       %
\usepackage{amsfonts}       %
\usepackage{nicefrac}       %
\usepackage{microtype}      %
\usepackage{colortbl}
\usepackage{multirow}

\usepackage{times}
\usepackage{latexsym}
\usepackage{graphicx}
\usepackage{calc}
\usepackage{booktabs}
\usepackage{array}
\usepackage{natbib}
\usepackage{enumerate}
\usepackage{stmaryrd}
\usepackage{caption}
\usepackage{wrapfig}
\usepackage{calc}
\usepackage{xfrac}
\usepackage{bm}
\usepackage{tabularx}

\newlength\myheight
\newlength\mydepth
\settototalheight\myheight{Xygp}
\usepackage[bottom]{footmisc}  %

\usepackage[usenames,dvipsnames]{xcolor}
\definecolor{lightgray}{gray}{0.6}
\definecolor{orange}{HTML}{f39c12}
\definecolor{blue}{HTML}{3498db}

\usepackage{amsmath,amsfonts,bm}

\def\eqref#1{equation~\ref{#1}}
\def\1{\bm{1}}

\def\vone{{\bm{1}}}

\def\va{{\bm{a}}}
\def\vb{{\bm{b}}}
\def\vc{{\bm{c}}}

\def\vm{{\bm{m}}}

\def\vv{{\bm{v}}}

\def\vx{{\bm{x}}}
\def\vy{{\bm{y}}}

\def\mA{{\bm{A}}}
\def\mB{{\bm{B}}}
\def\mC{{\bm{C}}}

\def\mH{{\bm{H}}}

\def\mK{{\bm{K}}}

\def\mM{{\bm{M}}}

\def\mQ{{\bm{Q}}}
\def\mR{{\bm{R}}}

\def\mV{{\bm{V}}}
\def\mW{{\bm{W}}}
\def\mX{{\bm{X}}}
\def\mY{{\bm{Y}}}

\DeclareMathAlphabet{\mathsfit}{\encodingdefault}{\sfdefault}{m}{sl}
\SetMathAlphabet{\mathsfit}{bold}{\encodingdefault}{\sfdefault}{bx}{n}
\newcommand{\tens}[1]{\bm{\mathsfit{#1}}}

\def\tG{{\tens{G}}}

\def\tT{{\tens{T}}}

\def\tW{{\tens{W}}}

\newcommand{\R}{\mathbb{R}}

\newcommand{\softmax}{\mathrm{softmax}}

\DeclareMathOperator*{\concat}{concat}
\DeclareMathOperator*{\stack}{stack}

\def\WK{\mW_{\!K}}
\def\WQ{\mW_{\!Q}}
\def\WV{\mW_{\!V}}
\def\tildeWK{\tilde{\mW}_{\!K}}
\def\tildeWQ{\tilde{\mW}_{\!Q}}

\newcommand{\codeurl}{\url{https://github.com/epfml/collaborative-attention}}
\usepackage{cleveref}

\usepackage[colorinlistoftodos,prependcaption,textsize=tiny,textwidth=3.5cm]{todonotes}

\title{Multi-Head Attention: \\ Collaborate Instead of Concatenate}
\author{Jean-Baptiste Cordonnier \and Andreas Loukas \and Martin Jaggi}
\date{\'Ecole Polytechnique F\'ed\'erale de Lausanne (EPFL)\\\vspace{.5em}\texttt{\{first.last\}@epfl.ch}}

\begin{document}

\maketitle

\begin{abstract}
  Attention layers are widely used in natural language processing (NLP) and are beginning to influence computer vision architectures.
  Training very large transformer models allowed significant improvement in both fields,
  but once trained, these networks show symptoms of over-parameterization.
  For instance, it is known that many attention heads can be pruned without impacting accuracy.
  This work aims to enhance current understanding on how multiple heads interact.
  Motivated by the observation that attention heads learn redundant key/query projections, we propose a collaborative multi-head attention layer that enables heads to learn shared projections.
  Our scheme decreases the number of parameters in an attention layer and can be used as a drop-in replacement in any transformer architecture.
  Our experiments confirm that sharing key/query dimensions can be exploited in language understanding, machine translation and vision.
  We also show that it is possible to re-parametrize a pre-trained multi-head attention layer into our collaborative attention layer.
  Collaborative multi-head attention reduces the size of the key and query projections by 4 for same accuracy and speed.
  Our code is public.\footnote{\codeurl}
\end{abstract}

\section{Introduction}

Since the invention of attention~\citep{bahdanau2014neural} and its popularization in the transformer architecture~\citep{vaswani2017attention},
multi-head attention (MHA) has become the de facto architecture for natural language understanding tasks \citep{bert} and neural machine translation.
Attention mechanisms have also gained traction in computer vision following the work of \citet{standalone}, and \citet{Bello_2019_ICCV}.
Nevertheless, despite their wide adoption, we currently lack solid theoretical understanding of how transformers operate. In fact, many of their modules and hyperparameters are derived from empirical evidence that are possibly circumstantial.

The uncertainty is amplified in multi-head attention, where both the roles and interactions between heads are still poorly understood.
Empirically, it is well known that using multiple heads can improve model accuracy.
However, not all heads are equally informative, and it has been shown that certain heads can be pruned without impacting model performance.
For instance, \citet{voita-etal-2019-analyzing} presented a method to quantify head utility and prune redundant members.
\citet{michel2019sixteenheads} questioned the utility of multiple heads by testing the effect of heavy pruning in several settings.
On the other hand, \citet{Cordonnier2020On} prove that multiple heads are needed for self-attention to perform convolution, specifically requiring one head per pixel in the filter's receptive field.

This work aims to better detect and quantify head redundancy by asking whether independent heads learn overlapping or distinct concepts.
We discover that many of the key/query projected dimensions are redundant, as trained concatenated heads tend to compute their attention patterns on common features.
Our finding implies that MHA can be re-parametrized with better weight sharing for these common projections and a lower number of parameters.

\pagebreak

Our \emph{contributions} are the following:
\begin{itemize}
  \setlength\itemsep{-.1em}
  \item We characterize the redundancy in the key/query features across heads using PCA in \Cref{ssec:pca}.
  \item This analysis leads to a natural re-parametrization of the concatenation-based multi-head attention layer. \Cref{sec:collab_head} describes the \emph{collaborative attention} letting heads learn common key and query projections.
  \item
  \Cref{sec:tensor_decomposition} describes how canonical tensor decomposition can be leveraged to
  reparametrize post-hoc any pre-trained transformers to use collaborative attention.
  \item We confirm through a wide range of experiments (\Cref{sec:experiments})
  that collaborative heads is an efficient multi-head scheme%
 . For instance in NMT when training from scratch, we reduce the number of parameters of the attention layers by 35\%
  \item As a side %
contribution, we identify a discrepancy between the theory and some implementations of attention layers and show that by correctly modeling the biases of key and query layers, we can clearly differentiate between context and content-based attention.
\end{itemize}

\paragraph{Related work.}
Some recent works have studied alternative head duplication schemes.
\citet{shazeer2020talkingheads} proposed to orchestrate collaboration between heads on top of the dot product attention scores.
In contrast, our approach increases heads expressivity by leveraging the low-rankness accross heads to share common query/key dimensions.
A practical approach to compress multi-head attention layer is to prune less informative heads \citep{voita-etal-2019-analyzing,michel2019sixteenheads} yielding significant reduction in number of parameters.
However, pruning still requires to pre-train the original model with all heads.
Rather than a post-hoc fix, we take a step toward understanding what is shared accross heads that makes heads redundant or degenerated.
Beyond the number of ``necessary'' heads, finding the adequate head dimension is also an open question.
\citet{bhojanapalli2020lowrank} found that the division of the key/query projection between heads gives rise to a low-rank bottleneck for each attention head expressivity that can be fixed by increasing the head sizes.
Finally, the tensor decomposition was previously used by \citet{kim2016compression} to compress trained CNN and factorize common convolutional filters.

\section{Multi-Head Attention}

We first review standard multi-head attention introduced by \cite{vaswani2017attention}.

\subsection{Attention}

Let $\mX~\in~\R^{T\times D_{\textit{in}}}$ and $\mY~\in~\R^{T'\times D_{\textit{in}}}$ be two input matrices consisting of respectively $T$ and $T'$ tokens of ${D_{\textit{in}}}$ dimensions each.
An attention layer maps each of the $T$ query token from $D_{\textit{in}}$ to $D_{\textit{out}}$ dimensions as follows:
\begin{align}
  \operatorname{Attention}(\mQ, \mK, \mV)
  =
  \softmax
  \left(
    \frac{\mQ \mK^\top}{\sqrt{d_k}}
  \right)
  \mV, \label{eq:attention}\\
  \text{with}\ \ \mQ = \mX \WQ,\, \mK = \mY \WK,\, \mV = \mY \WV
\end{align}
The layer is parametrized by a query matrix $\WQ~\in~\R^{D_{\textit{in}} \times D_{k}}$, a key matrix $\WK~\in~\R^{D_{\textit{in}} \times D_{k}}$ and a value matrix $\WV~\in~\R^{D_{\textit{in}} \times D_{\textit{out}}}$.
Using attention on the same sequence (i.e.~$\mX=\mY$) is known as self-attention and is the basic building block of the transformer architecture.

\subsection{Content vs. Context}

Some re-implementations of the original transformer architecture\footnote{For instance: BERT original implementation, HuggingFace re-implementation, FairSeq encoder-decoder transformer, Vision Transformer.} use biases in the linear layers.
This differs from the attention operator defined in \cref{eq:attention} where the biases $\vb_Q$ and $\vb_K \in \R^{D_k}$ are ommited.
Key and query projections are computed as $\mK = \mX\WK + \vone_{T \times 1}\vb_K$ and $\mQ = \mY\WQ + \vone_{T \times 1}\vb_Q$, respectively, where $\vone_{a\times b}$ is an all one matrix of dimension $a \times b$.
The exact computation of the (unscaled) attention scores can be decomposed as follows:
\begin{align}
  \mQ\mK^\top =& (\mX\WQ + \vone_{T \times 1}\vb_Q^\top)(\mY \WK + \vone_{T \times 1}\vb_K^\top)^\top\\
  =&
  \underbrace{\mX\WQ  \WK^\top \mY^\top}_{\text{context}}
  + \underbrace{\vone_{T\times 1} \vb_Q^\top \WK^\top \mY^\top}_{\text{content}}\nonumber\\
  &+ \mX \WQ \vb_K \vone_{1\times T}
  + \vone_{T \times T} \vb_Q^\top \vb_K \label{eq:decomp_conte_t}
\end{align}
As the last two terms of eq.~(\ref{eq:decomp_conte_t}) have a constant contribution over all entries of the same row, they do not contribute to the computed attention probabilities (softmax is shift invariant and $\softmax(\vx + c) = \softmax(\vx),\,\forall c$).
On the other hand, the first two terms have a clear meaning: $\mX\WQ  \WK^\top \mY^\top$ considers the relation between keys and query pairs, whereas $\vone_{T\times 1} \vb_Q^\top \WK^\top \mY^\top$ computes attention solely based on key content.
This decomposition suggests that the bias $\vb_K$ of the key layer can always be disabled without any consequence.

\subsection{Multi-Head Attention}

Traditionally, the attention mechanism is replicated by concatenation to obtain multi-head attention defined for $N_h$ heads as:
\begin{align}
  &\operatorname{MultiHead}(\mX, \mY) = \concat_{i \in {[N_h]}}\big[\mH^{(i)}\big] \; \mW_O\\
  &\mH^{(i)} = \operatorname{Attention}(\mX\WQ^{(i)}, \mY\WK^{(i)}, \mY\WV^{(i)}),
  \label{eq:multi-head-concat}
\end{align}
where distinct parameter matrices $\WQ^{(i)}, \WK^{(i)} \in \R^{D_{\textit{in}}\times d_k}$ and $\WV^{(i)} \in \R^{D_{\textit{in}}\times d_{\textit{out}}}$ are learned for each head $i \in [N_h]$ and the extra parameter matrix $\mW_{O}~\in~\R^{N_h d_{\textit{out}} \times D_{\textit{out}}}$ projects the concatenation of the $N_h$ head outputs (each in $\R^{d_{\textit{out}}}$) to the output space $\R^{D_{\textit{out}}}$.
In the multi-head setting, we call $d_k$ the dimension of each head and $D_k=N_hd_k$ the total dimension of the query/key space.

\section{Improving the Multi-Head Mechanism}
\label{sec:collab_head}

Head concatenation is a simple and remarkably practical setup that gives empirical improvements.
However, we show that another path could have been taken instead of concatenation.
As the multiple heads are inherently solving similar tasks, they can collaborate instead of being independent.

\begin{figure*}
  \centering
\begin{subfigure}{}
  \centering
  \includegraphics[width=.48\linewidth]{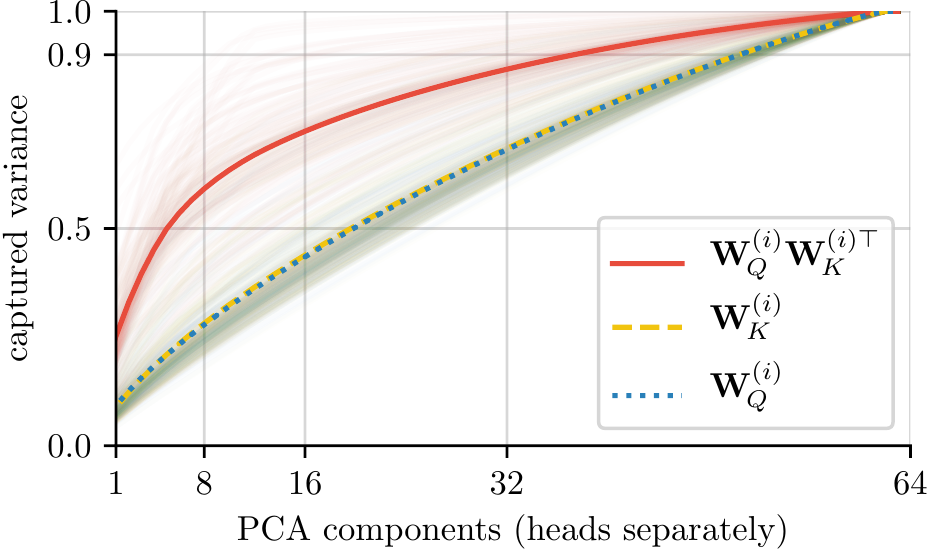}
\end{subfigure}%
\hfill%
\begin{subfigure}{}
  \centering
  \includegraphics[width=.48\linewidth]{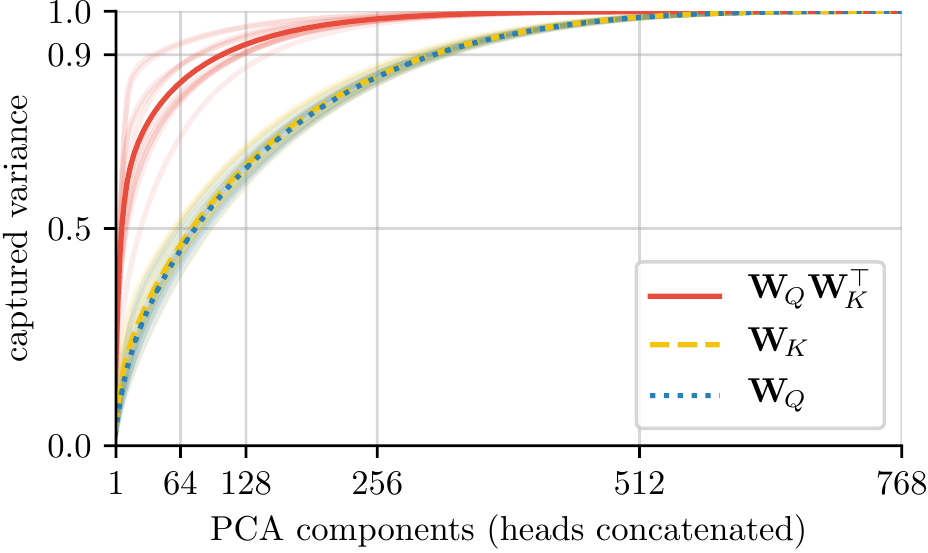}
\end{subfigure}
  \caption{
    Cumulative captured variance of the key query matrices per head separately (\emph{left}) and per layer with concatenated heads (\emph{right}).
    Matrices are taken from a pre-trained BERT-base model with $N_h = 12$ heads of dimension $d_k=64$.
    Bold lines show the means.
    Even though, by themselves, heads are not low rank (\emph{left}),
    the product of their concatenation $\WQ\WK^\top$ is low rank (\emph{right, in red}).
    Hence, the heads are sharing common projections in their column-space.
  }
  \label{fig:captured_variance}
\end{figure*}

\subsection{How much do heads have in common?}
\label{ssec:pca}

We hypothesize that some heads might attend on similar features in the input space, for example computing high attention on the verb of a sentence or extracting some dimensions of the positional encoding.
To verify this hypothesis, it does not suffice to look at the similarity between query (or key) matrices $\{\WQ^{(i)}\}_{i\in [N_h]}$ of different heads.
To illustrate this issue, consider the case where two heads are computing the same key/query representations up to a unitary matrix $\mR\in\R^{d_k \times d_k}$ such that
$\WQ^{(2)} = \WQ^{(1)} \mR$ and $\WK^{(2)} = \WK^{(1)} \mR$.

Even though the two heads are computing identical attention scores, i.e.~$\WQ^{(1)}\mR\mR^\top\WK^{(1)\top}=\WQ^{(1)} \WK^{(1)\top}$,
they can have orthogonal column-spaces and
the concatenation $[\WQ^{(1)}, \WQ^{(2)}]\in \R^{D_{\textit{in}}\times 2d_k}$ can be full rank.

To disregard artificial differences due to common rotations or scaling of the key/query spaces, we study the similarity of the product $\WQ^{(i)}\WK^{(i)\top} \in \R^{D_{\textit{in}} \times D_{\textit{in}}}$ across heads.
\Cref{fig:captured_variance} shows the captured energy by the principal components of the key, query matrices and their product.
It can be seen on the left that single head key/query matrices $\WQ^{(i)}\WK^{(i)\top}$ are not low rank on average.
However, as seen on the right, even if parameter matrices taken separately are not low rank, their concatenation is indeed low rank. This means that heads, though acting independently, learn to focus on the same subspaces.
The phenomenon is quite pronounced: \emph{one third of the dimensions suffices to capture almost all the energy of} $\WQ\WK^\top$, which suggests that there is inefficiency in the way multi-head attention currently operate.

\subsection{Collaborative Multi-Head Attention}

Following the observation that heads' key/query projections learn redundant projections,
we propose to learn key/query projections for all heads at once and to let each head use a re-weighting of these projections.
Our collaborative head attention is defined as follows:
\begin{align}
  &\hspace{-1em}\operatorname{CollabHead}(\mX, \mY) = \concat_{i \in {[N_h]}}\big[\mH^{(i)}\big] \; \mW_{\!O}\\
  &\hspace{-1em}\mH^{(i)} = \operatorname{Attention}(\mX \tildeWQ \operatorname{diag}(\vm_i), \mY \tildeWK, \mY\WV^{(i)})\,.
\end{align}
The main difference with standard multi-head attention defined in \cref{eq:multi-head-concat} is that we do not duplicate the key and query matrices for each head.
Instead, each head learns a mixing vector $\vm_i \in \R^{\tilde D_k}$ that defines a custom dot product over the $\tilde D_k$ projected dimensions of the shared matrices $\tildeWQ$ and $\tildeWK$ of dimension $D_{\textit{in}} \times \tilde D_k$.
This approach leads to:
\begin{enumerate}[(i)]
  \item adaptive head expressiveness, with heads being able to use more or fewer dimensions according to attention pattern complexity;
  \item more parameter efficient representations, as learned projections are shared between heads, hence stored and learned only once.
\end{enumerate}

\begin{figure*}[t]
  \centering
  \includegraphics[width=.7\linewidth]{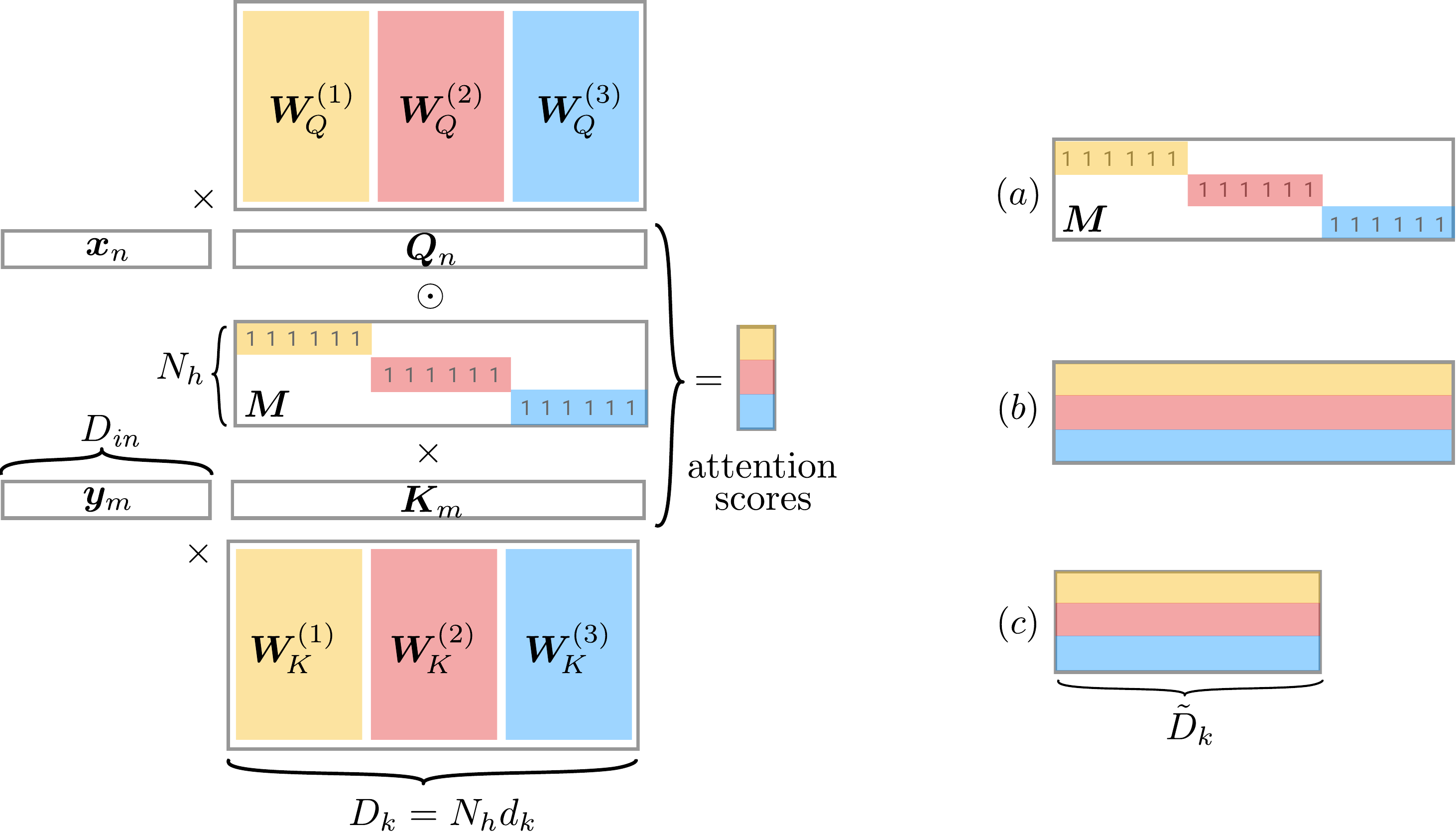}
  \caption{\emph{Left}: computation of the attention scores between tokens $\vx_n$ and $\vy_m$ using a standard concatenated multi-head attention with $N_h~\!=~\!3$ independent heads.
  The block structure of the mixing matrix $\mM$ enforces that each head dot products non overlapping dimensions.
  \emph{Right}: we propose to use more general mixing matrices $\mM$ than
  (a) heads concatenation, such as
  (b) sharing heads projections by learning all entries of the matrix;
  (c) compressing the number of projections from $D_k$ to $\tilde D_k$ as heads can share redundant projections.
  }
  \label{fig:attention_mask_tensor}
\end{figure*}

It is instructive to observe how standard multi-head attention (where heads are simply concatenated) can be seen as a special case of our collaborative framework (with $\tilde D_k = N_h d_k$).
The left of Figure~\ref{fig:attention_mask_tensor} displays the standard attention computed between $\vx_n$ and $\vy_m$ input vectors with the mixing matrix
\begin{align}
  \mM &:= \concat_{i \in [N_h]} \big[ \vm_i \big] \in \R^{N_h \times \tilde D_k}\,,
\end{align}
laying out the mixing vectors $\vm_i$ as rows.
In the concatenated MHA, the mixing vector $\vm_i$ for the $i$-th head is a vector with ones aligned with the $d_k$ dimensions allocated to the $i$-th head among the $D_k=N_h d_k$ total dimensions.

Some alternative collaborative schema can be seen on the right side of Figure~\ref{fig:attention_mask_tensor}.
By learning the mixing vectors $\{\vm_i\}_{i\in [N_h]}$ instead of fixing them to this ``blocks-of-1'' structure, we increase the expressive power of each head for a negligible increase in the number of parameters.
The size $d_k$ of each head, arbitrarily set to 64 in most implementations, is now adaptive and the heads can attend to a smaller or bigger subspace if needed.

\subsection{Head Collaboration as Tensor Decomposition}
\label{sec:tensor_decomposition}

As we show next, there is a simple way to convert any standard attention layer to collaborative attention without retraining.
To this end, we must extract the common dimensions between query/key matrices $\{\WQ^{(i)} \WK^{(i)\top} \in \R^{D_{\textit{in}}\times D_{\textit{in}}}\}_{i\in [N_h]}$ across the different heads.
This can be solved using the Tucker tensor decomposition \citep{tucker} of the 3rd-order tensor
\begin{align}
  \tW_{\!Q\!K} :=
  \stack_{i\in[N_h]} \left[ \mW_Q^{(i)}\mW_K^{(i)\top} \right] \in \R^{N_h \times D_{\textit{in}} \times D_{\textit{in}}}\,.
\end{align}

Following the notation\footnote{$\circ$ represents the vector outer product} of \citet{kolda2009tensor}, the Tucker decomposition of a tensor $\tT \in \R^{I\times J \times K}$ is written as
\begin{align}
  \tT \approx \tG \times_1 \mA \times_2 \mB \times_3 \mC = \sum_{p=1}^P \sum_{q=1}^Q \sum_{r=1}^R g_{pqr} \, \va_p \circ \vb_q \circ \vc_r
\end{align}
with $\mA \in \R^{I\times P}$, $\mB \in \R^{J\times Q}$, and $\mC\in \R^{K \times R}$ being factor matrices, whereas $\tG \in \R^{P\times Q \times R}$ is the core tensor. Intuitively, the core entry $g_{pqr} = \tG_{p,q,r}$ quantifies the level of interaction between the components $\va_p,  \vb_q$, and $\vc_r$.

In the case of attention, it suffices to consider the dot product of the aligned key/query components of the $\mQ$ and $\mK$ matrices, which means that the core tensor is super-diagonal (i.e. $g_{pqr} \not= 0$ only if $q=r$).
We further simplify the Tucker decomposition by setting the factors dimensions $P, Q$ and  $R$ to $\tilde D_k$,
a single interpretable hyperparameter equal to the dimension of the \emph{shared} key/query space that controls the amount of compression of the decomposition into collaborative heads.
These changes lead to a special case of Tucker decomposition called the canonical decomposition, also known as CP or PARAFAC \citep{Hars1970,kolda2009tensor}. For any positive rank $R$, the decomposition yields:
\begin{align}
  \tT \approx \sum_{r=1}^R \va_r \circ \vb_r \circ \vc_r =: \big\llbracket \mA, \mB, \mC \big\rrbracket\,,
  \label{eq:canonical_decomposition}
\end{align}
with $\mA\in \R^{I\times R}$, $\mB\in \R^{J\times R}$ and $\mC \in \R^{K\times R}$.

What is remarkable is that the above can be used to express any (trained) attention layer parametrized by $\{\WQ^{(i)}, \vb_Q^{(i)}, \WK^{(i)}, \vb_K^{(i)}\}_{i\in[N_h]}$ as a collaborative layer.
In particular, if we apply the decomposition to the stacked heads $\tW_{\!Q\!K}$ we obtain the three matrices
$
  \llbracket \mM, \tildeWQ , \tildeWK \rrbracket %
$
that define a collaborative attention layer: the mixing matrix $\mM \in \R^{N_h \times \tilde D_{k}}$, as well as the key and query projection matrices $\tildeWQ$, $\tildeWK \in \R^{D_{\textit{in}}\times \tilde D_k}$.

On the other hand, biases can be easily dealt with based on the content/context decomposition of \cref{eq:decomp_conte_t}, by storing for each head the vector
\begin{align}
  \vv_i &= \WK^{(i)} \vb_Q^{(i)} \in \R^{D_{\textit{in}}}.
\end{align}

With this in place, the computation of the (unscaled) attention score for the $i$-th head is given by:
\begin{align}
  \hspace{-4mm}\left(\mX\WQ^{(i)} + \vone_{T\times 1}\vb_Q^\top \right)
  \left(\mY\WK^{(i)} + \vone_{T\times 1}\vb_K^\top \right)^\top
  \nonumber\\\approx
  \mX \tildeWQ \operatorname{diag}(\vm_i) \tildeWK^\top \mY^\top
  +
  \vone_{T \times 1} \vv_i^\top \mY^\top,
\end{align}
where $\vm_i$ is the $i$-th row of $\mM$.
If $\tilde D_k \geq D_k$ the decomposition is exact (\cref{eq:canonical_decomposition} is an equality) and our collaborative heads layer can express any concatenation-based attention layer.
We also note that the proposed re-parametrization can be applied to the attention layers of a wide variety transformer architectures,
such as the ones proposed by~\citet{bert,distilbert,Lan2020ALBERT}.

\subsection{Parameter and Computation Efficiency}

\newcolumntype{R}{>{\raggedleft}X}
\begin{table}
\centering
\caption{Comparison of a layer of concatenate vs. collaborative MHA with chosen total key dimension $D_k$ for $T=128$ tokens and batch size 32 on a V100 GPU.}
\label{tab:param_gain}
\setlength{\tabcolsep}{4pt}
\begin{tabular}{@{}rr@{\hskip .3em}>{\color{black!50} \scriptsize}lr@{\hskip .3em}>{\color{black!50} \scriptsize}lr@{\hskip .3em}>{\color{black!50} \scriptsize}lr@{\hskip .3em}>{\color{black!50} \scriptsize}l@{}}
\toprule
   & \multicolumn{2}{c}{Params} & \multicolumn{2}{c}{FLOPS} & \multicolumn{2}{c}{Train {\footnotesize(ms)}} & \multicolumn{2}{c}{Infer {\footnotesize(ms)}} \\
   \midrule
 \multicolumn{3}{@{}l@{}}{\textbf{Concat}} \\
 768 & 2.36M & $\times 1.00$ & 10.9G & $\times 1.00$ & 5.27 & $\times 1.00$ & 2.28 & $\times 1.00$ \\
 \midrule
 \multicolumn{3}{@{}l@{}}{\textbf{Collab}} \\
 64  & 1.29M & $\times 0.55$ &  8.9G & $\times 0.82$ & 4.04 & $\times 0.77$ & 1.75 & $\times 0.77$ \\
 128 & 1.39M & $\times 0.59$ & 12.1G & $\times 1.11$ & 4.37 & $\times 0.83$ & 1.79 & $\times 0.79$ \\
 256 & 1.59M & $\times 0.67$ & 18.5G & $\times 1.70$ & 5.40 & $\times 1.02$ & 2.11 & $\times 0.93$ \\
 384 & 1.78M & $\times 0.76$ & 25.0G & $\times 2.29$ & 6.27 & $\times 1.19$ & 2.51 & $\times 0.99$ \\
\bottomrule
\end{tabular}%
\end{table}

Collaborative MHA introduces weight sharing across the key/query projections and decreases the number of parameters.
In contrast to standard attention layers where the size of the heads is set to $d_k$ and the key/query layers project into a space of dimension $D_k=N_h d_k$, our shared key/query dimension $\tilde D_k$ of collaborative MHA can be set freely.
We show below that that swapping from standard to collaborative heads results in a linear decrease in the total number of parameters of the transformer.

\paragraph{Parameters.}
The proposed change of the architecture replaces the key and query matrices by three matrices of adjustable dimensions.
Collaborative attention uses
$(2 D_{\textit{in}} + N_h ) \tilde D_k$ parameters,
as compared to
$2 D_{\textit{in}} D_k$ in the standard case (ignoring biases).
Hence, the compression ratio is $\approx D_k / \tilde D_k$, controlled by the shared key dimension~$\tilde D_k$.
The collaborative factorization introduces a new matrix $\mM$ of dimension $N_h \times \tilde D_k$.
Nevertheless, as the number of heads is small compared to the hidden dimension (in BERT-base $N_h = 12$ whereas $D_{\textit{in}} = 768$), the extra parameter matrix yields only a negligible increase as compared to the bigger size of the query/key/values matrices of dimension $D_{\textit{in}} \times D_k$.

\paragraph{Computational cost.}
To compute the attention scores between $T$ tokens for all the $N_h$ heads, collaborative MHA requires
$T(2D_{in} + N_h)\tilde D_k + T^2 N_h\tilde D_k$  FLOPS, while traditional concatenation-based MHA
uses $2T D_{in} D_k + T^2 N_h d_k$ FLOPS.
\Cref{tab:param_gain} shows that in practice both layers have similar speed at inference for reasonable compression $\tilde D_k = D_k / 2$.

\paragraph{Comparison with head pruning.}
Head pruning gives a linear decrease in parameters of the attention layers as it also affect the $\mW_V$ and $\mW_O$ matrices.
The parameter gains of our method are not as significant as we leave these matrices untouched and compression cannot be as drastic as head pruning \citep{michel2019sixteenheads}.
However, pruning still requires to pre-train the original model with all heads.
Whereas at this point the compression benefits are not on par with pruning, our approach presents a principled/elegant way to reduce inefficiencies rather than relying on post-hoc processing.

\section{Experiments}
\label{sec:experiments}

We present experiments on transformers applied to diverse applications: Neural Machine Translation (NMT) on WMT14 EN-DE, Natural Language Understanding (NLU) on the GLUE benchmark \citep{wang-etal-2018-glue} and image classification on ImageNet \citep{ILSVRC15}.
We first show that collaborative MHA is a drop-in replacement for concatenation-based MHA in the transformer architecture.
Training a transformer with collaborative attention from scratch on NMT (\S\ref{ssec:nmt}) and vision tasks (\S\ref{ssec:vit}) allows to reduce the number of attention parameters to achieve the same performance or improve it.
As pretraining transformers is computationally expensive, a significant part of our investigations are then run on pretrained models (\S\ref{ssec:reparam_vit},\S\ref{ssec:nlu}).
We display that even post-hoc re-parametrization with tensor decomposition is effective on many transformer architectures in NLU %
and vision. %
This confirms our observation that some of the key/query projections are redundant.

\paragraph{Setup.} NMT experiments are based on the FairSeq \citep{ott2019fairseq} implementation of transformer-base by \cite{vaswani2017attention}.
For the NLU experiments, we implemented the collaborative MHA layer as an extension of the Transformers library \citep{Wolf2019HuggingFacesTS}.
The flexibility of our layer allows it to be applied to most of the existing transformer architectures, either at pre-training or after fine-tuning using tensor decomposition.
Our image classification experiments are based on DeiT \citep{touvron2021training} implementation of Vision Transformers \citep{dosovitskiy2021an}.
We use the tensor decomposition library Tensorly \citep{tensorly} with the PyTorch backend \citep{paszke2017automatic} to reparameterize pre-trained attention layers.
Our code and datasets are publicly available\footnote{\codeurl} and all hyperparameters are specified in the Appendix.

\subsection{Collaborative MHA for NMT}
\label{ssec:nmt}

We replace the concatenation-based MHA layers of an encoder-decoder transformer by our collaborative MHA and evaluate it on the WMT14 English-to-German translation task.
Following \citep{vaswani2017attention}, we train on the WMT16 train corpus, apply checkpoint averaging and report compound split tokenized BLEU.
We use the same hyperparameters as the baseline for all our runs.
Results are shown in \Cref{fig:nmt_blue}.
Our run of the original base transformer with $N_h=8$ heads and $D_k=512$ key/query total dimensions achieves 27.40 BLUE (27.30 in the original paper).

As observed by \cite{vaswani2017attention}, decreasing the key/query head size $d_k$ degrades the performance ({\color{orange} $\bm{\times}$} in \Cref{fig:nmt_blue}).
However, with collaborative heads ({\color{blue} +} in \Cref{fig:nmt_blue}), the shared key/query dimension can be reduced by 4$\times$ without decreasing the BLEU score.
As feed-forward layers and embeddings are left untouched, this translates to a 35\%
When setting a total key/query dimension of $D_k=64$, corresponding to $d_k = 8$ dimensions per head, the classic MHA model suffers a drop of 0.6 BLEU points, meanwhile the collaborative MHA stays within 0.1 point of the transformer-base model using concatenation.

We conclude that sharing key/query projections across heads allows  attention features to be learned and stored only once.
This weight sharing enables decreasing~$D_k$ without sacrificing expressiveness.

\subsection{Collaborative MHA for vision}
\label{ssec:vit}

\begin{minipage}{\linewidth}
\centering
\captionof{table}{Comparison of the BLEU score on the WMT14 EN-DE translation task for an encoder-decoder transformer-base \citep{vaswani2017attention} using collaborate vs. concatenate heads with key/query dimension $D_k$. Collaborative attention consistently improves the BLEU score, $D_k$ can be decreased from 512 to 128 without any drop in performance.} 
\label{tab:nmt_blue}
  \begin{tabular}{@{}lc@{\hspace{.5em}}cc@{\hspace{.5em}}cc@{\hspace{.5em}}c@{}}
\toprule
     & \multicolumn{2}{c}{BLEU $\shortuparrow$}  & \multicolumn{2}{c}{params ($\times10^6$)} &  \multicolumn{2}{c}{time (h)} \\
     \cmidrule(lr){2-3}  \cmidrule(lr){4-5} \cmidrule(l){6-7}
 $D_k$   & concat. & collab. & concat. & collab. & concat. & collab. \\
\midrule
 512 & 27.40  & 27.58  & 60.9            & 61.0   & 18.0   & 21.0   \\
 256 & 27.10  & 27.41  & 56.2            & 56.2   & 17.3   & 19.0   \\
 128 & 26.89  & 27.40  & 53.8            & 53.8   & 17.3   & 18.4   \\
 64  & 26.77  & 27.31  & 52.6            & 52.7   & 16.9   & 17.9   \\
\bottomrule
\end{tabular}%
\vspace{2em}
\includegraphics[width=.45\linewidth]{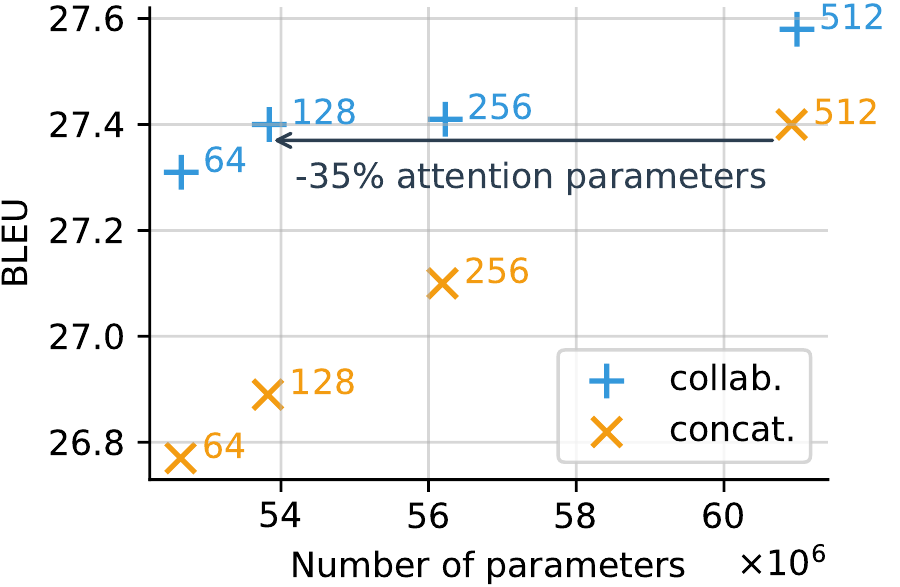}%
\hspace{.05\linewidth}%
\includegraphics[width=.45\linewidth]{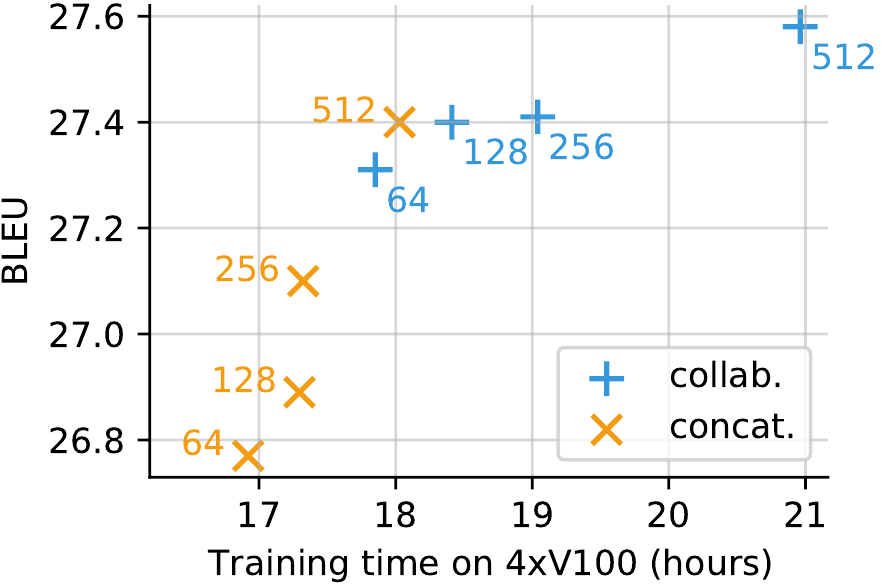}
\vspace{.5em}
\captionof{figure}{Comparison of BLEU score on WMT14 EN-DE translation task presented in \Cref{tab:nmt_blue}.
We visualize performance as a function of number of parameters (\emph{left}) and training time (\emph{right}).
}
\label{fig:nmt_blue}
\end{minipage}

We apply collaborative attention to the image modality motivated by the recent advances of transformers in computer vision.
We train a vision transformers~\citep{Cordonnier2020On,dosovitskiy2021an} on ImageNet~\citep{ILSVRC15} at resolution 224$\times$224 following the training
procedure from DeiT~\cite{touvron2021training}. As a single run of DeiT-B requires 16 V100 GPUs for approximately 4 days, we chose a smaller model to  train a vision transformer from scratch with limited computing ressources in mind.
We disregarded DeiT-Tiny~\cite{touvron2021training} as it downscales the number of heads (12 to 3) and hidden dimension (768 to 192) which deceives the purpose of measuring heads sharing exploited by collaborative attention.
We run instead a shallower DeiT-B model with depth 3 (called DeiT-B3) instead of 12 for a proxy of the performance of the full DeiT-B model.
A run takes approximately 3 days on 4 V100 GPUs following the training procedure of DeiT-Tiny.

The results when training from scratch are presented in \Cref{tab:vit_scratch}.
First, we note that decreasing the key/query dimension to $D_k=384$ does not alter performance even with concatenation-based attention but Acc\@1 drops by 0.6\%
Moreover, using colaborative heads consistantly improves the model performance over concatenation-based attention with the same key/query dimension and matches DeiT-B3 with 4 times less dimensions.
Specifically, using collaborative attention with $D_k = 384$ and 192 matches the performance of the baseline with a gain in total number of parameters of 8\%

\begin{table}[t]
\centering
\caption{Performance on ImageNet dev. of a 3-layers DeiT-B \citep{touvron2021training} trained from scratch.}
\label{tab:vit_scratch}
\begin{tabular}{@{}lcccr@{\hskip .3em}>{\color{black!50} \scriptsize}l@{}}
\toprule
  Model &$D_k$         & Acc@1 & Acc@5  &\multicolumn{2}{c}{Params} \\
  \midrule
  DeiT-B3        & 768 &  68.1 & 87.5 & 22.8M & $\times 1.00$ \\
  DeiT-B3        & 384 &  68.1 & 87.7 & 21.0M & $\times 0.92$ \\
  DeiT-B3        & 192 &  67.5 & 87.2 & 20.2M & $\times 0.89$ \\
  \hline
  DeiT-B3 collab & 384 &  68.2 & 87.8 & 21.0M & $\times 0.92$ \\
  DeiT-B3 collab & 192 &  68.0 & 87.8 & 20.2M & $\times 0.89$ \\
\bottomrule
\end{tabular}%
\end{table}

\subsection{Re-parametrization for vision}
\label{ssec:reparam_vit}
To further investigate the performance of DeiT-B with collaborative heads, we re-parametrize pre-trained vision transformers.
Starting from a pretrained DeiT-B model,
we apply the Tucker tensor decomposition (\S\ref{sec:tensor_decomposition}) to swap all the attention layers with collaborative attention for different shared key/query dimensions $D_k$.
This operation takes less than 10 minutes on a single GPU and no fine-tuning on the original data is needed.
The results in \Cref{tab:vit_reparam} show that compressing from $D_k=768$ to $512$ only alter Acc@1 by 0.1\%
A stronger compression to $D_k=256$ alters the Acc@1 on ImageNet by 1\%

\begin{table}[h]
\centering
\caption{Performance on ImageNet dev. of a pretrained DeiT-B \citep{touvron2021training} reparametrized with collaborative attention for different compression shared key dimension $D_k$.}
\label{tab:vit_reparam}
\begin{tabular}{@{}lcccr@{\hskip .3em}>{\color{black!50} \scriptsize}l@{}}
\toprule
  Model &$D_k$ & Acc@1 & Acc@5  &\multicolumn{2}{c}{Params} \\
  \midrule
  DeiT-B concat                & 768           & 81.7 & 95.6 & 86.5M & $\times 1.00$ \\
  \midrule
  DeiT-B collab                & 768           & 81.8 & 95.6 & 86.8M & $\times 1.00$\\
  {\footnotesize (reparam.)}   & 512           & 81.6 & 95.5 & 82.0M & $\times 0.95$\\
                               & 384           & 81.3 & 95.4 & 79.6M & $\times 0.92$\\
                               & 256           & 80.7 & 95.1 & 77.3M & $\times 0.89$\\
\bottomrule
\end{tabular}%
\end{table}

\subsection{Re-parametrization for NLU}
\label{ssec:nlu}

\begin{table*}
  \centering
  \caption{Performance of collaborative MHA on the GLUE benchmark \citep{wang-etal-2018-glue}.
  We report the median of 3 runs for BERT \citep{bert}, DistilBERT \citep{distilbert} and ALBERT \citep{Lan2020ALBERT} with collaborative heads and different compression controlled by $\tilde D_k$.
  Comparing the original models ($D_k=768$) with their compressed counter part shows that the number of parameters can be decreased with less than 1.5\%
  }
  \label{table:glue_results}
  \resizebox{\linewidth}{!}{%
  \small
  \begin{tabular}{lccccccccccc}
    \toprule
    Model              & $\tilde D_k$ &    params    & CoLA & SST-2 & MRPC      & STS-B     & QQP       & MNLI      & QNLI & RTE  & \textbf{Avg.}\\
    \midrule
    \textbf{BERT-base}
     &  - & 108.3M & 54.7 & 91.7  & 88.8/83.8 & 88.8/88.7 & 87.6/90.8 & 84.1      & 90.9 & 63.2 &   83.0 \\
          &  768   & 108.5M & 56.8 & 90.1  & 89.6/85.1 & 89.2/88.9 & 86.8/90.2 & 83.4      & 90.2 & 65.3 &  83.2 \\
                       & 384   & 101.4M  & 56.3 & 90.7  & 87.7/82.4 & 88.3/88.0 & 86.3/90.0 & 83.0      & 90.1 & 65.3 &  82.5 \\
\rowcolor{black!10}
                       & 256   & \phantom{1}99.0M & 52.6 & 90.1  & 88.1/82.6 & 87.5/87.2 & 85.9/89.6 & 82.7      & 89.5 & 62.5 &  81.7 \\
                       & 128   & \phantom{1}96.6M & 43.5 & 89.5  & 83.4/75.2 & 84.5/84.3 & 81.1/85.8 & 79.4      & 86.7 & 60.7 &  77.6 \\

    \midrule
    \textbf{DistilBERT} & - &   \phantom{1}66.4M  &          46.6 & 89.8 & 87.0/82.1 & 84.0/83.7 & 86.2/89.8 & 81.9 & 88.1 & 60.3 & 80.0  \\
        \rowcolor{black!10}
                    & 384 & \phantom{1}62.9M &  45.6 & 89.2 & 86.6/80.9 & 81.7/81.9 & 86.1/89.6 & 81.1 & 87.0 & 60.7 & 79.1  \\

    \midrule
    \textbf{ALBERT} & - & \phantom{1}11.7M
                                             & 58.3 & 90.7 & 90.8/87.5 & 91.2/90.8 & 87.5/90.7  & 85.2       & 91.7 &   73.7 &  85.3 \\
    \rowcolor{black!10}
                    & 512 & \phantom{1}11.3M  & 51.1 & 86.0 & 91.4/88.0 & 88.6/88.2 & 87.2/90.4 & 84.2      & 90.2 &  69.0  & 83.1  \\
                    & 384 & \phantom{1}11.1M   & 40.7 & 89.6 & 82.3/71.1 & 86.0/85.6 & 87.2/90.5 & 84.4      & 90.0 &  49.5 & 77.9
                    \\

    \bottomrule
  \end{tabular}%
  }
\end{table*}

\begin{figure}[b]
  \centering
  \includegraphics[width=.4\linewidth]{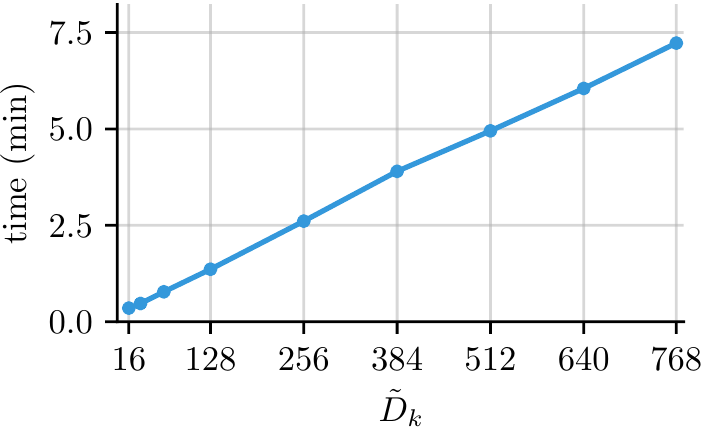}
  {\footnotesize \caption{Time to decompose BERT-base from $D_k=768$ to $\tilde D_k$.}
  \label{fig:decomposition_time}}
\end{figure}

\begin{figure*}[t]
  \centering
\begin{subfigure}{}
  \centering
  \includegraphics[width=.3\linewidth]{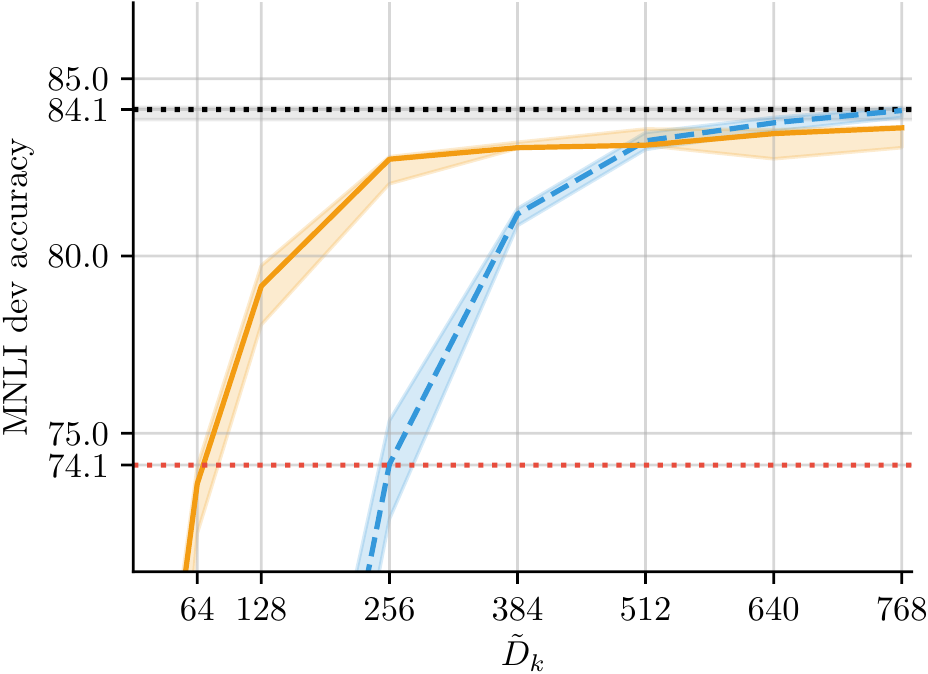}
\end{subfigure}%
\hspace{.9em}%
\begin{subfigure}{}
  \centering
  \includegraphics[width=.3\linewidth]{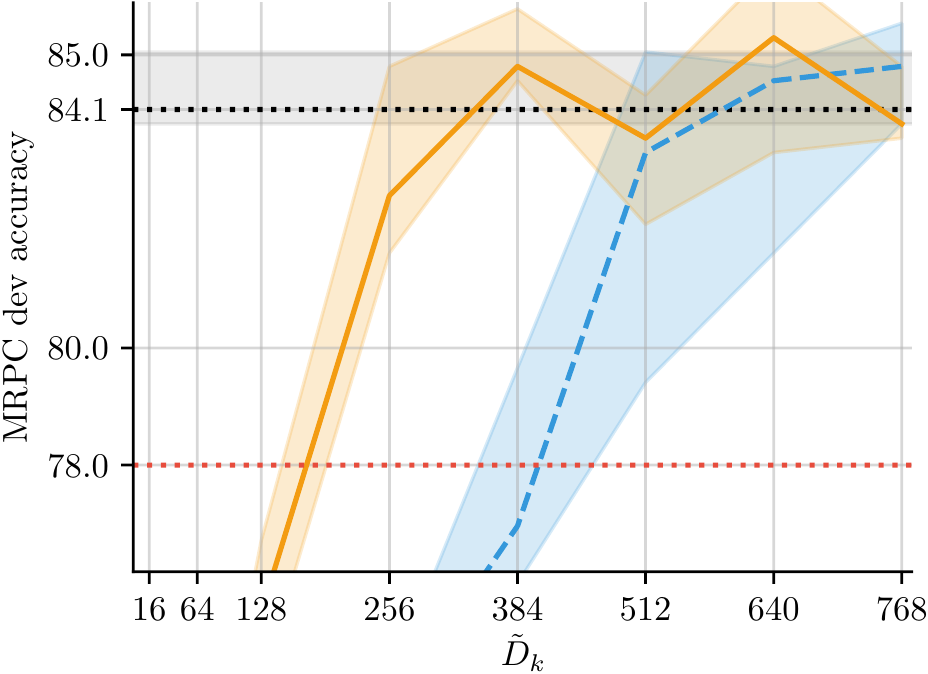}
\end{subfigure}%
\hspace{.9em}%
\begin{subfigure}{}
  \centering
  \includegraphics[width=.3\linewidth]{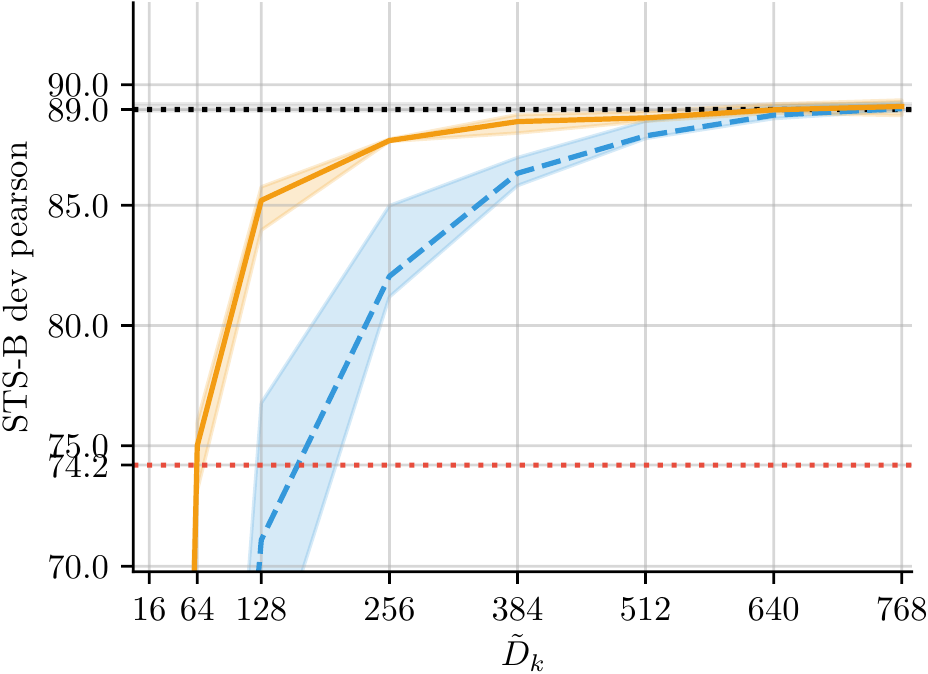}
\end{subfigure}%
\hspace{.9em}%
  \caption{
  Performance on MNLI, MRPC and STS-B datasets
  of a \protect\includegraphics[height=\myheight]{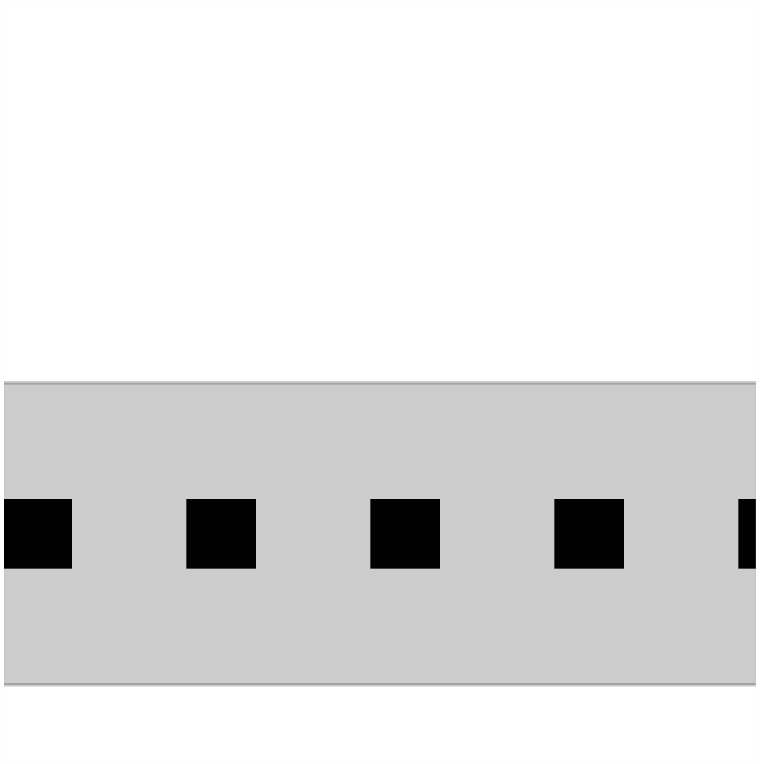}~fine-tuned BERT-base model,
   \protect\includegraphics[height=\myheight]{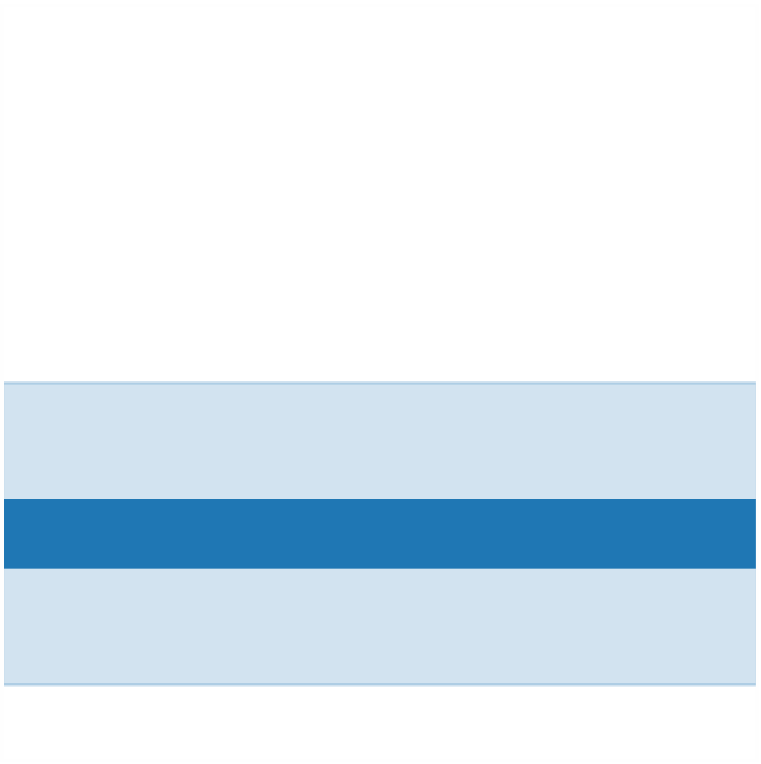}~decomposed with collaborative heads of compressed dimension $\tilde D_k$ (\emph{horizontal axis}).
  \protect\includegraphics[height=\myheight]{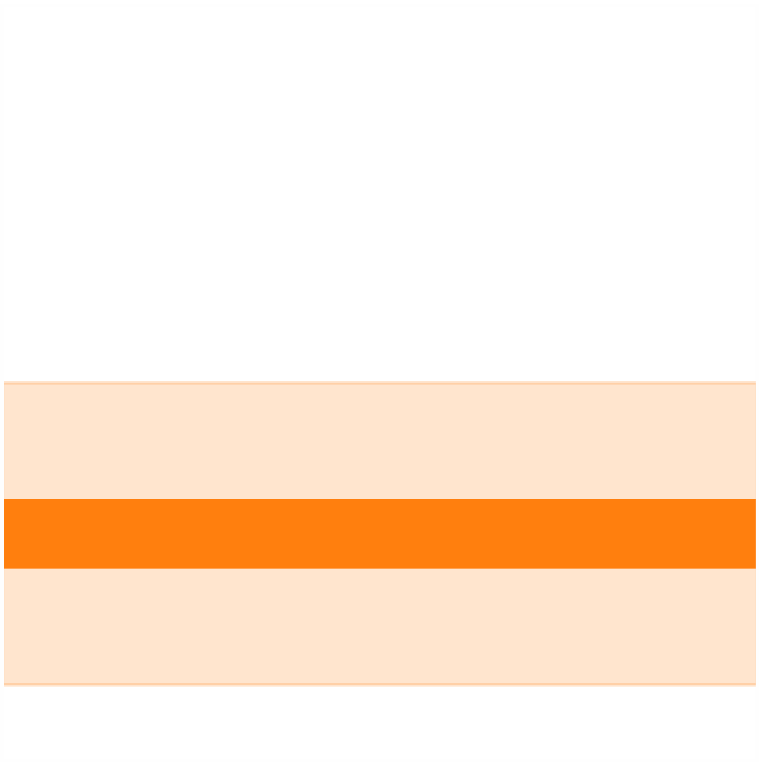}~Repeating fine-tuning after compression can make the model recover the original performance when compression was drastic.
  We report \protect\includegraphics[height=\myheight]{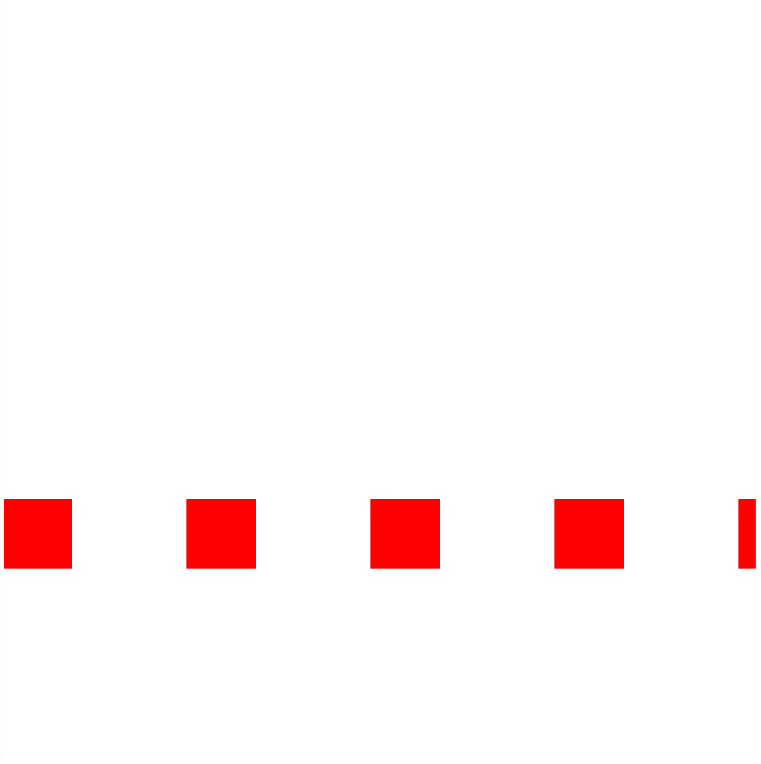}~the GLUE baseline for reference.
  }
  \label{fig:Dk_mrpc}
\end{figure*}

We turn to experiments on Natural Language Understanding (NLU) tasks, where transformers have been decisive in improving the state-of-the-art.
As pre-training on large text corpora remains an expensive task, we leverage the post-hoc re-parametrization introduced in \Cref{sec:tensor_decomposition} to cast already pre-trained models into their collaborative form. %
We proceed in 3 steps for each GLUE task \citep{wang-etal-2018-glue}.
First, we take a pre-trained transformer and fine-tune it on each task individually.
Secondly, we replace all the attention layers by our collaborative MHA using tensor decomposition to compute $\tildeWQ$, $\tildeWK$ and $\mM$ and re-parametrize the biases into $\vv$. This step only takes a few minutes as shown in \Cref{fig:decomposition_time}.
Finally, we fine-tune the compressed model again and evaluate its performance.

We experiment with a pre-trained BERT-base model \citep{bert}.
We also repurpose two variants of BERT designed to be more parameter efficient: ALBERT \citep{Lan2020ALBERT}, an improved transformer with a single layer unrolled, and DistilBERT \citep{distilbert} a smaller version of BERT trained with distillation.
We report in \Cref{table:glue_results} the median performance of 3 independent runs of the models on the GLUE benchmark \citep{wang-etal-2018-glue}.

We first verify that tensor decomposition without compression ($\tilde D_k = D_k = 768$) does not alter performance. As shown in Table~\ref{table:glue_results}, both BERT-base and its decomposition performs similarly with an average score of 83.0\%

We then experiment with compressed decomposition using a smaller~$\tilde D_k$.
Comparing the original models with their well-performing compressed counterpart (gray rows) shows that the key/query dimension of BERT and DistilBERT can be reduced by 2$\times$ and~3$\times$ respectively without sacrificing more than 1.5\%
This is especially remarkable given that DistilBERT was designed to be a parameter-efficient version of BERT.
It seems that ALBERT suffers more from compression, but the dimension can be reduced by a factor 1.5$\times$ with minor performance degradation.
We suspect that unrolling the same attention layer over the depth of the transformer forces the heads to use different projections and decreases their overlap, decreasing the opportunity for weight-sharing.
Our hypothesis is that better performance may be obtained by pre-training the whole BERT architecture variants from scratch.

\paragraph{Recovering from compression with fine-tuning.}
We further investigate the necessity of the second fine-tuning---step 3 of our experimental protocol---after the model compression.
\Cref{fig:Dk_mrpc} shows the performance of BERT-base on 3 GLUE tasks for different compression parameters $\tilde D_k$ with and without the second fine-tuning.
We find that for compression up to 1.5$\times$ (from $D_k = 768$ to $\tilde D_k =512$), the re-parametrization is accurate and performance is maintained without fine-tuning again.
Further compressing the model starts to affect performance.
Nevertheless, for compression by up to 3$\times$ (to $\tilde D_k = 256$), this loss can readily be recovered by a second fine-tuning (\emph{in orange}).

\section{Conclusion}

This work showed that concatenated heads in multi-head attention models tend to learn redundant query/key representations.
To mitigate this issue, we propose to replace concatenation-based MHA by collaborative MHA in standard transformers. When our layer is used instead of standard MHA in NMT and vision, it enables to divide the key/query dimension by 4 without performance drop.

Our model can impact every transformer architecture and our codebase also provides post-hoc compression of trained networks.
We believe training collaborative MHA from scratch can help heads to extract meaningful shared query/key features.
Our approach presents a principled and elegant way to reduce inefficiencies rather than relying on post-hoc processing such as head pruning.
We hope that our findings are a first step towards an efficient multi-head mechanism that does not result in degenerated or redundant heads.%

\newpage

\nocite{wandb}
\nocite{tange_ole_2018_1146014}

\section*{Acknowledgments}

We acknowledge the support of Google Cloud for computational credits. Jean-Baptiste Cordonnier is
thankful to the Swiss Data Science Center (SDSC) for funding this work. Andreas Loukas would
like to thank the Swiss National Science Foundation for supporting him in the context of the project
``Deep Learning for Graph-Structured Data'' (grant number PZ00P2 179981).

\bibliography{refs}
\bibliographystyle{icml2021}

\newpage
\appendix

\onecolumn

\begin{center}
{\huge \textbf{Supplementary Material}}
\end{center}

\section{Hyperparameters for Neural Machine Translation Experiments}

Our implementation is based on Fairseq implementation \cite{ott2019fairseq}. We report in the following tables the specification of the architecture. We used the default hyperparameters if they are not specified below.

\begin{table}[h]
\centering
\begin{tabular}{@{}ll@{}}
  \toprule
  \multicolumn{2}{c}{Transformer architecture parameters}            \\
  \midrule
  dataset & wmt16\_en\_de\_bpe32k\\
  architecture & transformer\_wmt\_en\_de\\
  layers & 6\\
  heads & 8\\
  hidden-dim & 512\\
  collaborative-heads & "encoder\_cross\_decoder" or "none"\\
  key-dim & 64, 128, 256, 512\\
  share-all-embeddings & True\\
  optimizer & adam\\
  adam-betas & (0.9, 0.98)\\
  clip-norm & 0.0\\
  lr & 0.0007\\
  min-lr & 1e-09\\
  lr-scheduler & inverse\_sqrt\\
  warmup-updates & 4000\\
  warmup-init-lr & 1e-07\\
  dropout & 0.1\\
  weight-decay & 0.0\\
  criterion & label\_smoothed\_cross\_entropy\\
  label-smoothing & 0.1\\
  max-tokens & 3584\\
  update-freq & 2\\
  fp16 & True\\
    \bottomrule
\end{tabular}
  \caption{Hyperparameters for the NMT experiment.}
\end{table}

\section{Hyperparameters for ImageNet Experiments}

Our code is based on DeiT \url{https://github.com/facebookresearch/deit}.
We report the hyperparameters used in \Cref{tab:comp_hyperparameters}.

\begin{table}[h]
\centering
\scalebox{0.9}
{
\begin{tabular}{lccc}
\toprule
Methods & DeiT-B3 \\
\midrule
Epochs   & 300     \\
\midrule
Batch size & 1024\\
Optimizer & AdamW\\
     learning rate       &   $0.0005\times \frac{\textrm{batchsize}}{512} $  \\
     Learning rate decay  & cosine  \\
     Weight decay         & 0.05    \\
     Warmup epochs   & 5       \\
\midrule
     Label smoothing $\epsilon$ & 0.1     \\
     Dropout      & \xmark     \\
     Stoch. Depth &  0.1 \\
     Repeated Aug & \cmark \\
     Gradient Clip. & \xmark \\
\midrule
     Rand Augment          & 9/0.5 \\
     Mixup prob.   & 0.8     \\
     Cutmix prob.   & 1.0    \\
     Erasing prob.  & 0.25    \\
 \bottomrule
\end{tabular}}
\vspace{-8pt}
\caption{
 Optimization hyper-parameters to train DeiT-B3.
\label{tab:comp_hyperparameters}}
\end{table}

\section{Hyperparameters for Natural Language Understanding Experiments}

We use standard models downloadable from HuggingFace repository along with their configuration presented in \Cref{table:models_params_nlu}.
We use HuggingFace default hyperparameters for GLUE fine-tuning specified in \Cref{table:hparams_nlu}.
We train with a learning rate of $2\cdot10^{-5}$ for 3 epochs for all datasets except SST-2 and RTE where we train for 10 epochs.
In preliminary experiments, we tried to tune the tensor decomposition tolerance hyperparameter among $\{10^{-6}, 10^{-7}, 10^{-8}\}$ but did not see significant improvement and kept the default $10^{-6}$ for all our experiments.

\begin{table}[h]
  \centering
  \begin{tabular}{@{}lll@{}}
    \toprule
    Models      &        & \\
    \midrule
    BERT-base & \cite{bert} & \texttt{bert-base-cased}\\
    DistilBERT & \cite{distilbert} & \texttt{distilbert-base-cased}\\
    ALBERT & \cite{Lan2020ALBERT} & \texttt{albert-base-v2}\\
    \bottomrule
  \end{tabular}
  \caption{Model references.}
  \label{table:models_params_nlu}
\end{table}

\begin{table}[h]
  \centering
  \begin{tabular}{@{}ll@{}}
    \toprule
    \multicolumn{2}{c}{GLUE fine-tuning hyperparameters} \\
    \midrule
    Number of epochs & 3 for all tasks but 10 for SST-2 and RTE \\
    Batch size & 32 \\
    Learning rate & 2e-5\\
    Adam $\epsilon$ & 1e-8 \\
    Max gradient norm & 1 \\
    Weight decay & 0\\
    Decomposition tolerance & 1e-6\\
    \bottomrule
  \end{tabular}
  \caption{Optimization hyperparameters.}
  \label{table:hparams_nlu}
\end{table}

\end{document}